\documentclass{bmvc2k}

\title{Annealing Genetic GAN for Minority Oversampling}

\addauthor{Jingyu Hao}{haojy5@mail2.sysu.edu.cn}{1}
\addauthor{Chengjia Wang}{chengjia.wang@ed.ac.uk}{2}
\addauthor{Heye Zhang}{zhangheye@mail.sysu.edu.cn}{1}
\addauthor{Guang Yang}{g.yang@imperial.ac.uk}{3}
\addinstitution{
 School of Biomedical Engineering\\
 Sun Yat-sen University\\
 Shenzhen, CHN
}
\addinstitution{
 University of Edinburgh\\
 London, UK
}
\addinstitution{
 Imperial College London\\
 London, UK 
}
\runninghead{Jingyu Hao}{AGGAN for Minority Oversampling}

\def\eg{\emph{e.g}\bmvaOneDot}

\def\etal{\emph{et al}\bmvaOneDot}
\usepackage{amsmath}
\usepackage{amsthm}
\usepackage{booktabs}
\usepackage{algorithm}
\usepackage{algorithmic}
\usepackage{bm}
\usepackage{graphicx}
\usepackage{multirow}
\newtheorem{corollary}{Corollary}

\begin{document}

\maketitle

\begin{abstract}
The key to overcome class imbalance problems is to capture the distribution of minority class accurately. Generative Adversarial Networks (GANs) have shown some potentials to tackle class imbalance problems due to their capability of reproducing data distributions given ample training data samples. However, the scarce samples of one or more classes still pose a great challenge for GANs to learn accurate distributions for the minority classes. In this work, we propose an Annealing Genetic GAN (AGGAN) method, which aims to reproduce the distributions closest to the ones of the minority classes using only limited data samples. Our AGGAN renovates the training of GANs as an evolutionary process that incorporates the mechanism of simulated annealing. In particular, the generator uses different training strategies to generate multiple offspring and retain the best. Then, we use the Metropolis criterion in the simulated annealing to decide whether we should update the best offspring for the generator. As the Metropolis criterion allows a certain chance to accept the worse solutions, it enables our AGGAN steering away from the local optimum. According to both theoretical analysis and experimental studies on multiple imbalanced image datasets, we prove that the proposed training strategy can enable our AGGAN to reproduce the distributions of minority classes from scarce samples and provide an effective and robust solution for the class imbalance problem.
\end{abstract}

%-------------------------------------------------------------------------
\section{Introduction}

In machine learning applications, class imbalance problem, i.e., differences in prior class probabilities, hinder the performance of many standard classifiers \cite{GalarA}. The key to improve the classification performance with imbalanced data is to capture the distributions of minority classes precisely \cite{he2009learning}. However, it is difficult to learn accurate distributions from scarce samples of the minority classes. A common strategy to tackle class imbalance problem is to increase the minority samples in order to have a better representation for the distributions of the minority classes. In particular, previous methods usually increased the size of minority classes by replicating or interpolating the samples from the minority classes. However, repetition may cause the problem of over-fitting because the samples from minority classes are overemphasised. On the other hand, because data are normally sitting in a high dimensional space, interpolation is nontrivial and may generate low-quality samples due to the complexity of the data manifold \cite{ando2017deep}.

\begin{figure}[!t]
\centering
\includegraphics[width=5in]{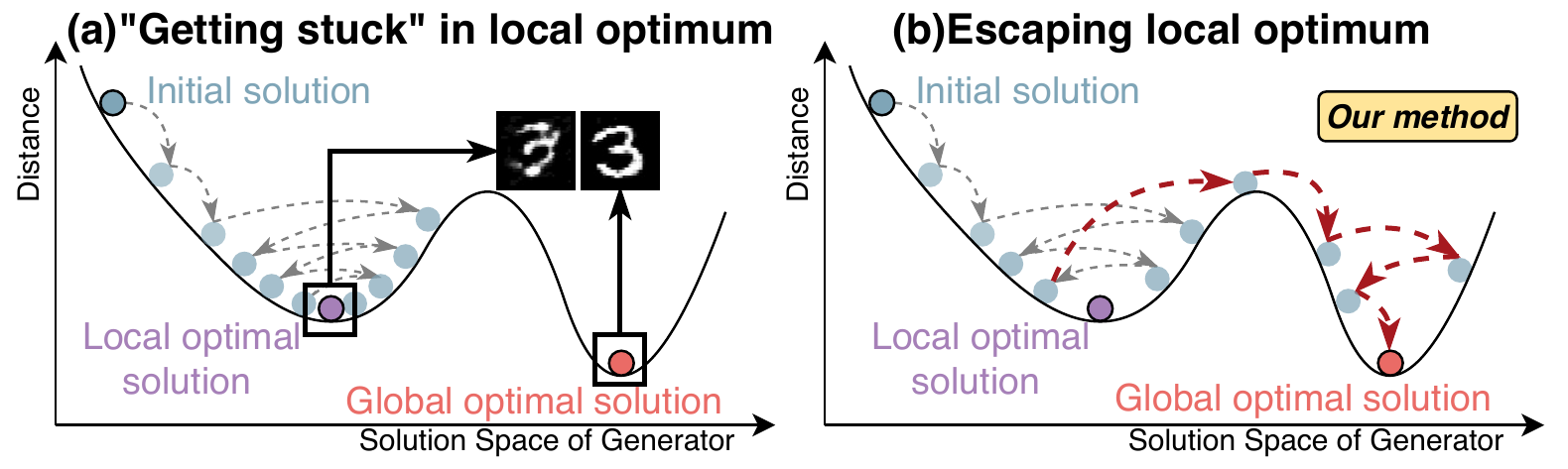}
\caption{(a) Schematic illustration of local optimum trapping for GANs. In this situation, GANs will keep fluctuating around the local optimum until the end of training and cannot learn a distribution that is close enough to the minority class. Consequently, the samples GANs generated may have low quality. (b) Our method integrates the simulated annealing genetic algorithm into the training of GANs. In doing so, GANs may update to a worse solution with a decreasing probability, which enables GANs to escape from the local optimum.}
\label{fig_sim1}
\end{figure}

Recently, Generative Adversarial Networks (GANs) have shown some potentials to tackle class imbalance problems because theoretically they are able to reproduce the distributions of minority classes through adversarial learning. In the training process of GANs, the generator learns the mapping from a latent encoding space to the minority class distribution, and the discriminator needs to determine whether an input sample is actually drawn from the minority class or created by the generator \cite{goodfellow2014generative}. As these two networks confront each other constantly, the performances of the generator and discriminator are improved alternately. Finally, the generator reproduces the distributions of the minority classes that the discriminator would not be able to distinguish from the actual minority classes.

There are many successful applications of using GANs. However, GANs can easily get stuck at local optimum when they try to learn the distributions from scarce samples of the minority classes that is also known as mode collapse \cite{Arjovsky2017Towards} as shown in Figure \ref{fig_sim1} (a). A more effective training strategy is highly in demand for GANs to avoid the trapping at the local optimum. Previous attempts tried to avoid local optimum by improving adversarial learning objectives \cite{mao2017least} \cite{ZhaoEnergy} \cite{arjovsky2017wasserstein}. However, these previous strategies still used a single adversarial learning target in the training process that might still fail to fully overcome the local optimum problem and might also have other limitations, e.g., Wasserstein distance has non-convergent limit cycles near the equilibrium \cite{nagarajan2017gradient}.

Considering the problems and limitations of those previously proposed methods, we propose a new model, namely Annealing Genetic GAN (AGGAN), to incorporate simulated annealing genetic algorithm into the training process of GANs to avoid the local optimum trapping problem (Figure \ref{fig_sim1} (b)). The primary contributions of the proposed method are summarised as follows:

\begin{itemize}
\item We develop a strategy to incorporate simulated annealing genetic algorithm into the training process of GANs to avoid the local optimum trapping.
\item Through theoretical analysis, we prove that the simulated annealing genetic algorithm enables our proposed AGGAN to reproduce the distributions closest to the minority classes.
\item We also conduct comprehensive experimental studies and show that our proposed AGGAN method can solve the class imbalance problem efficiently and effectively.
\end{itemize}

\section{Related Work}

\paragraph{Class Imbalance Problem}
Class imbalance is a common problem in practical classification tasks. The scarce samples of a minority class make it difficult for the classifier to find the boundaries between the distributions of different classes correctly. Therefore, the key to solve class imbalance is to learn an accurate distribution of the minority classes with limited samples. Oversampling can improve the accuracy of the learning on the distributions of minority classes by increasing minority samples, which is a common method to tackle the class imbalance problem \cite{he2009learning}. Random oversampling, Synthetic Minority Over-Sampling Technique (SMOTE) \cite{chawla2002smote} and Border-line SMOTE\cite{han2005borderline} are commonly used oversampling methods in classic imbalance problems.
%Among various oversampling methods, random oversampling was the easiest to implement because it added samples by simply replicating them. However, the samples obtained using this strategy might not be generalised, which could lead to over-fitting. Interpolation is another way of oversampling, e.g., using Synthetic Minority Over-Sampling Technique (SMOTE) \cite{chawla2002smote}. The SMOTE could alleviate the phenomenon of over-fitting, but it assumed that all kinds of samples were surrounded by the samples of minority class and had strong blindness \cite{han2005borderline}. Border-line SMOTE was then developed based on the SMOTE, in which only the samples of minority class near the borderline were over-sampled. It could mitigate marginality and blindness problems of the SMOTE. 
However, when dealing with data in high dimensional space, the quality of the synthesised new data points could still be compromised due to noise and poor distance measurement in the high dimensional space \cite{lusa2013smote}.

\paragraph{Generative Adversarial Networks}
Recently, GANs have achieved great success in image generation \cite{zhang2017stackgan}\cite{Mao_2019_CVPR}, image-to-image synthesis \cite{isola2017image}, image super-resolution \cite{ledig2017photo}  and other applications due to its excellent capability of learning the data distributions by providing abundant training samples. In addition, GANs have also shown some potentials to solve class imbalance problems by learning the distributions of minority classes. Although GANs have been successfully applied in many tasks, due to the limited number of samples in the minority class, GANs may only be able to learn part of the minority class distribution at the end of training, and therefore could be trapped by the local optimum. Some studies have been done to enable GANs to learn a more accurate distribution during training process by utilising improved adversarial learning objectives (e.g. LSGAN \cite{mao2017least}, energy-based GAN \cite{ZhaoEnergy} and WGAN \cite{arjovsky2017wasserstein}. Nevertheless, there still exist limitations when using fixed adversarial training objectives in the training of GANs. %For example, since WGAN measured the Wasserstein distance between the generated distribution and the data distribution, it had to enforce the Lipschitz constraint on the discriminator, which might result in difficulties for optimisation \cite{nagarajan2017gradient}. 
More recently, Evolutionary-GAN (E-GAN) \cite{wang2019evolutionary} was proposed and multiple generators were created by different adversarial objectives to overcome the limitations of the fix adversarial objectives, and always kept the well-performed generator in the training process. However, local optimum trapping problem has yet been addressed.

%\paragraph{Simulated Annealing Genetic Algorithm}
%Genetic Algorithms (GA) and Simulated Annealing (SA) are the main methods for solving search and optimisation problems in high dimensional spaces \cite{Adler1993Genetic}. Recently, they have been introduced to optimise deep learning hyper-parameters and design deep network architectures. Galar et al. \cite{GalarA} proposed to use Cartesian genetic programming to designing convolutional neural network (CNN) architectures. Rere et al. \cite{RereSimulated} used SA to improve the performance of CNN, as an alternative approach for optimal deep learning using modern optimisation technique. Tao et al. \cite{tao2019variational} introduced a novel variational annealing strategy to stabilise and facilitate general training of GANs.

\section{Annealing Genetic GAN (AGGAN)}

\begin{figure}
\centering
\includegraphics[width=3.3in]{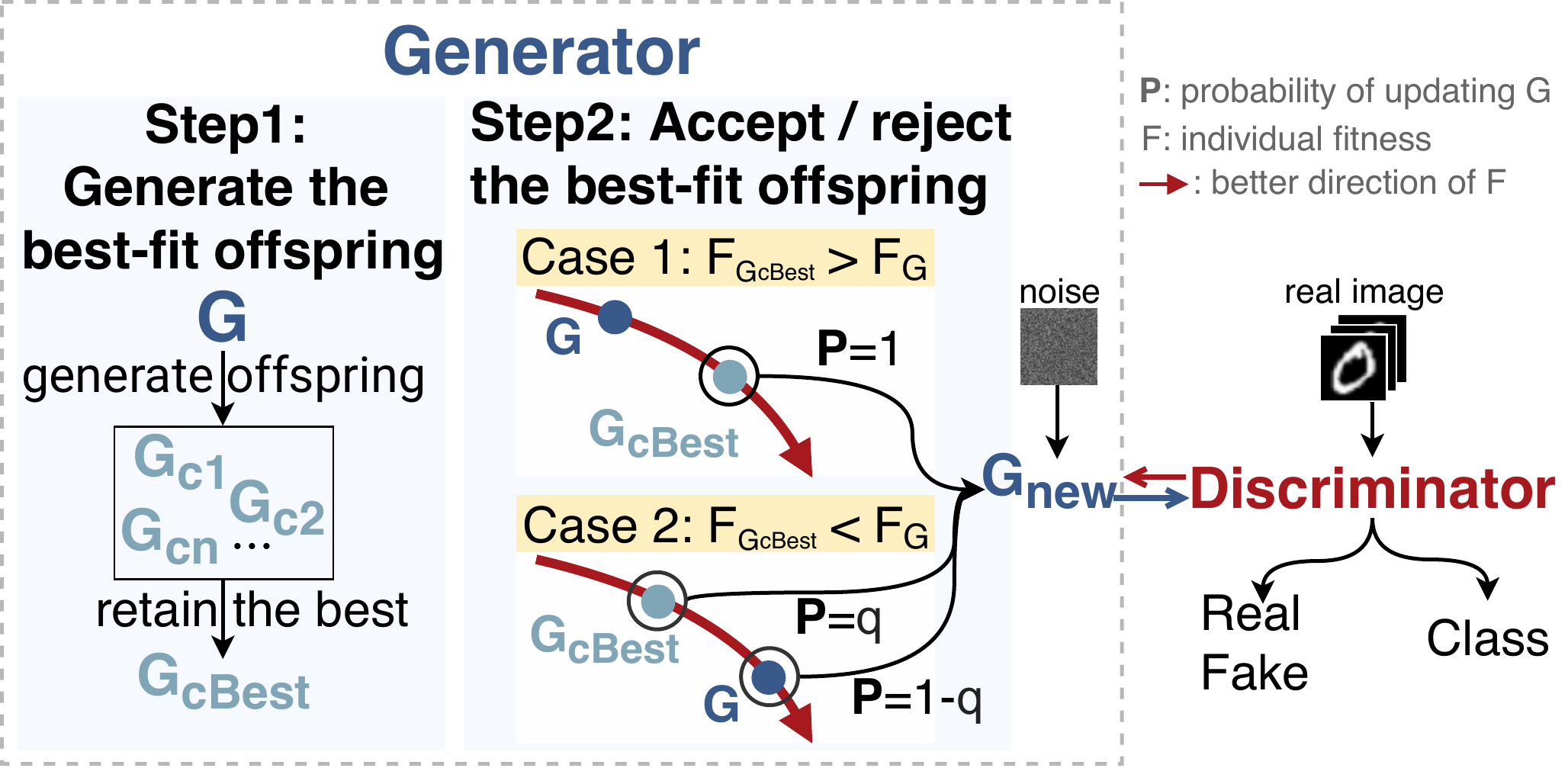}
\caption{We illustrate the training process of the generator here. This training process can be divided into two steps, which are described in Section 3 with more details.}
%\caption{The part inside the dashed box represents the training of the generator, which can be divided into two steps. The first step is to generate the best-fit offspring. The parent generator $G$ generates $G_{c1},G_{c2}...G_{cn}$ through different adversarial learning objectives. Assessing the individual fitness of all offspring, the offspring $G_{cbest}$ with the highest individual fitness is retained. The second step is to update the generator. We use the simulated annealing strategy to decide whether to update $G_{cbest}$ to the next generator $G_{new}$. If the individual fitness of $G_{cbest}$ is higher than the parent $G$, then we update $G_{cbest}$ to $G_{new}$ with a probability of 1. If the individual fitness of $G_{cbest}$ is lower than the parent $G$, $G_{cbest}$ is updated to $G_{new}$ with a probability of $P=q$, where $q$ is based on the temperature, annealing coefficient and the difference between the two individual fitness. This probability $q$ will continue to decrease with the training so that GANs can converge to the global optimum. After the training of the generator is finished, the samples from $G_{new}$ are input into the discriminator $D$ for training together with the real samples.}
\label{fig_sim2}
\end{figure}

In this paper, we propose AGGAN that aims to learn the accurate distribution from the minority class. First, our AGGAN uses different adversarial learning objectives to improve performance of the generator. Second, our AGGAN incorporates the mechanism of simulated annealing into the training so that the model can converge to the distribution that is closest to the minority class. In particular, in lieu of normal training of GANs, our AGGAN is trained as an evolutionary process, with the discriminator $D$ as the \it{environment}\rm~and the generator $G$ as the \it{individual}\rm. Each iteration of the individual is divided into two steps including (1) generating the best-fit offspring and (2) updating the generator as illustrated in Figure \ref{fig_sim2}.

\subsection{Generating the Best-Fit Offspring}
In each iteration, $G$ gives birth to different offspring $G_c$ by various adversarial learning objectives. Each $G_c$ represents a solution in the parameter space of the generator network. The individual fitness of offspring $G_{c}$ is evaluated based on the diversity and quality of the generated samples. Then the best-fit offspring $G_{cBest}$ is retained while other offspring are eliminated. This process of generating the best-fit offspring $G_{cBest}$ reflects the concept of `survival of the fittest' in the GA. The strategy of using different adversarial objectives overcomes the limitations of using a fixed objective and helps the final learned generator to achieve a better performance.

\subsection{Updating the Generator}
In our study, we propose to use the mechanism of SA to update the generator. If the individual fitness of $G_{cBest}$ was higher than $G$, $G_{cBest}$ will be updated to $G_{new}$ with a probability of 1. If the individual fitness of $G_{cBest}$ was lower than the previous generation $G$, $G_{cBest}$ will be updated to $G_{new}$ with a probability of $P$. The probability $P$ is determined based on the current temperature $T_c$ and the difference between the two individual fitness. The temperature $T_c$ gradually decreases from the initial temperature $T$ conditioning on the annealing coefficient $\alpha$. In doing so, updating $G$ with a decreasing probability in a worse direction enables AGGAN to asymptotically converge to the global optimum.

Finally, after updating the individual, the environment (i.e., the discriminator) $D$ is updated and the training loop of our AGGAN starts the next evolutionary iteration. As the training progresses, the data generated by G gradually close to the true distribution, which helps D to continuously improve classification accuracy.%Algorithm \ref{alg:algorithm} summarises the whole training procedure of the AGGAN.

\section{Theoretical Analysis}

In this section, we perform a theoretical analysis of the proposed AGGAN. We will prove that incorporating simulated annealing genetic algorithm into the training process of GANs can reinforce our AGGAN to learn the distribution closest to the minority class, that is, the AGGAN can converge to the global optimum solution with a probability of 1.

For elaboration purpose, we consider the training of GANs as a combinatorial optimisation problem. We consider this combinatorial optimisation problem as a pair of $(G,f)$, where $G$ is a finite set of the solution space of the generator $g$ and $f$ is the object function. The aim is to find a global optimum to minimise $f$. It is of note that the finiteness of $G$ implies that $f$ has at least one minimum over $G$.

Below we provide the definitions of generating the best-fit offspring and updating the generator in AGGAN from a mathematical perspective and prove that the simulated annealing genetic algorithm will make AGGAN to converge to the optimum solution as the training progressing.

\subsection{Definition}
\begin{table}[t]
	\centering
	\begin{tabular}{ll|ll}
		\hline
		Symbol & Definition & Symbol & Definition \\
		\hline
		$\bm{g}$ &Individual & $\bm{g_c}$ &Offspring of $g$ \\
		$\bm{g_{cbest}}$ &The best offspring of $g$ & $\bm{g_b}$ &The best $g$ so far \\
		$\bm{G}$ &Solution space of $g$ & $\bm{G_n}$ &Solution space for the $n_{th}$ iteration\\
	    $\bm{\{G_n|n\in N\}_g}$ & \multicolumn{3}{l}{The solution space sequence  with initial state g}\\
	 \hline
	\end{tabular}
	\caption{Definition of symbols.}
	\label{tab:plain}
\end{table}
For readability, we first define the symbols we used in Table \ref{tab:plain}. Then we give the mathematical definition of `generating the best-fit offspring' and `updating the generator' in AGGAN. In addition, in order to obtain the monotonicity without changing the training mechanism, we add a second element $g_b$ to each individual, which represents the best $g$ so far.

\paragraph{Generating the best-fit offspring \boldmath{$F_{gen}$}} 

First, we use a choice function to find the parent $g$ from $[g,g_b]$. Then $g$ generates offspring $g_c$ under the production function. Finally, we retain the best-fit offspring $g_{cbest}$ from an individual fitness function. All the steps above describe the whole process of `generating the best-fit offspring'. Since the choice function, production function and individual fitness function will not be mentioned in the analysis below, we collectively denote the above process using $F_{gen}$
\begin{align}
F_{gen}(g,g_{b})=g_{cbest}.
\end{align}

As the parameters of the choice function, production function and individual fitness function do not depend on the number of iterations, $F_{gen}$ will follow the same distribution with different number of iterations.

\paragraph{Updating the generator \boldmath{$F_{upd}$}}

The process of `updating the generator' uses the mechanism of SA. We define it as follows
\begin{equation}
 F_{upd}(g,g_{b},g_{cbest})=
 \left\{
 \begin{split}
 &[g,g'_{b}],\ \ \ \ \ \ \ \mbox{if}\ e^{-\frac{\Delta}{T}}<\gamma\\
 &[g_{cbest},g'_{b}], \mbox{otherwise}
 \end{split}
 \right.
 \end{equation}
 where 
 \begin{equation}
 g'_b=
 \left\{
 \begin{split}
 &g_b ,&\ \mbox{if}\ f(g_{cbest})>f(g_b)\\
 &g_{cbest} ,&\ \mbox{if}\ f(g_b)>f(g_{cbest})
 \end{split}
 \right.
 \end{equation}
and $\Delta$ is the difference between the individual fitness of $g$ and $g_{cbest}$, $\gamma$ is a random variable between 0 and 1, and $T$ denotes the temperature parameter for our AGGAN.

We combine $F_{upd}$ and $F_{gen}$ to obtain $F$ for the representation of the iterative process of the generator
\begin{align}\label{eq:1}
F(g,g_{b})=F_{upd}(g,g_{b},F_{gen}(g,g_{b})),
\end{align}
and we use $F_{n}$ to denote the $F$ in the $n_{th}$ iteration.

\subsection{Proof of Convergence}

In this section, according to Corollary \ref{th1}, we will prove that $\{G_n|n\in N\}_g$ satisfies the following properties, which can ensure that our AGGAN will converge to the global optimum with a probability of 1.
\begin{itemize}
	\item[1] \textbf{Monotonicity} The minimum value of $G_n$ on $f$ decreases as $n$ becomes larger.
	\item[2] \textbf{Homogeneity} If $\{G_n|n\in N\}_g$ is a Markov chain, and the $F_{n}$ have the same distribution then the chain is homogeneous.
\end{itemize}
%\noindent\textbf{Monotonicity}

First, we discuss the monotonicity of $\{G_n|n\in N\}_g$. Since $F_{upd}$ always keeps the best $f$ value, for any $g$ we have
\begin{align}\label{eq:2}
 min\{f(s)|s\in g\}\geq& min\{f(s)|s\in [g,g_b,g_{cbest}]\}\\
 \geq &min\{f(s)|s\in f_{upd}(g,g_b,g_{cbest})\}\nonumber
\end{align} holds.
%due to the conservativity of $f_s$. 

Hence by Eq. \ref{eq:1}, Eq. \ref{eq:2} and 
\begin{align}
 G_{n+1}= F_{n}(G_n,g_b)
\end{align} we have that 
\begin{align}
 mi&n\{f(s)|s \in G_{n+1}\}= min\{f(s)|s \in F_{n}(G_n,g_b)\}\nonumber\\
 &= min\{f(s)|s \in F_{upd}(G_n,g_b, F_{gen}(G_n,g_b))\}\nonumber\\
 &\leq min\{f(s)|s \in G_n\},
\end{align}
so that $\{G_n|n\in N\}_g$ is monotone.

Then we discuss the homogeneity. Due to the memory-less property of GANs, for any $n$, the conditional probability distribution of $G_{n+1}$ (conditional on both past and present states) depends only upon the $G_{n}$, not on the sequence of $G$ that preceded it. Then we have
\begin{align}
 &P(G_{n+1}=g_{n+1}|(G_n=g_n) \wedge ... \wedge (G_0=g))\nonumber\\
 %=&P(F_{n}(g_n)=g_{n+1}|[F_{n-1}(g_{n-1})=g_n] \wedge ...\nonumber\\
 % &... \wedge [F_{0}(g_0)=g_1])\nonumber\\
 %=&P(F_{n}(g_n)=g_{n+1})\\
 =&P(G_{n+1}=g_{n+1}|G_n=g_n),
\end{align}
so that $\{G_n|n\in N\}_g$ has Markov property. Since $F_{n}$ have the same distribution then
\begin{align}
 P(G_m=y|G_{m-1}=z)=P(G_n=y|G_{n-1}=z)
\end{align} 
for any $y,z\in G$ and $m,n\in N$. So that $\{G_n|n\in N\}_g$ is homogeneous.

The following Corollary \ref{th1} is derived from the Theorem 2 in the paper of Aarts\etal \cite{aarts1989general}.

\begin{corollary}\label{th1}
\it{Let $g\in G$ and the following conditions be satisfied:

(a) $\{G_n|n\in N\}_g$ is monotone

(b) $\{G_n|n\in N\}_g$ is homogeneous

(c) for every $h\in succ(g)$ there exists at least one accessible optimum.

Then $\{G_n|n\in N\}_g$ surely reaches an optimum.}
\end{corollary}

As shown above, we can prove that our AGGAN can converge to the global optimum, which is the closest solution to the minority class distribution with a probability of 1.

\section{Experimental Studies and Discussion}

We have used a collection of 6 image datasets for our
experiments, namely MNIST \cite{lecun1998gradient}, Fashion-MNIST \cite{xiao2017fashion}, SVHN \cite{netzer2011reading}, CIFAR-10 \cite{krizhevsky2009learning}, CelebA\cite{liu2015faceattributes}, and LSUN\cite{yu2015lsun}.
We evaluate our method in two imbalanced environments: two-class and multi-class.
In binary classification, because all the selected datasets are multi-class datasets, we randomly select two classes from each dataset. Choose Digit 5  (positive) and Digit 6 (negative) from MNIST, Sandal (positive) and Sneaker (negative) from Fashion-MNIST, and Airplane (positive) and Automobile (negative) from CIFAR-10, Digit 8 (positive) and Digit 9 (negative) from SVHN,  Eyeglasses (positive) and No-eyeglasses (negative) from CelebA, and choose Church(positive) and Classroom (negative) from LSUN.
In multi-classification, we transform the original balanced training in the same way as the paper of Mullick \etal \cite{mullick2019generative}.
And we define the Imbalance Ratio (IR) as
the number of training samples in the largest class divided by the smallest one.
%Because all the chosen datasets are multiclass datasets, we choose two similar classes from each dataset. Digit 3 (positive) and Digit 5 (negative) are chosen from the MNIST, T-shirt (positive) and dress (negative) are chosen from the Fashion-MNIST, ship (positive) and airplane (negative) are chosen from the CIFAR-10, and Digit 8 (positive) and Digit 9 (negative) are chosen from the SVHN. For all four datasets, we randomly select a disparate number of samples from each class to create imbalanced training sets, e.g., with an imbalance ratio of 1:100 (positive samples vs. negative samples).

\subsection{Implementation Details}
In the binary experiment, we have compared AGGAN, against baseline classifier network (CN), Os+CN (training set is random oversampled), ACGAN\cite{odena2017conditional} and Evolutionary-GAN\cite{wang2019evolutionary} (the version of the discriminator with classifier) to prove the effectiveness of our method. The same network structures are used for these different methods to achieve a fair comparison. In particular, our AGGAN and E-GAN use the same adversarial learning objectives to generate multiple offspring generators in the experiment, including the minimax loss, the modified minimax loss and the least-squares loss. The same evaluation function has been used to measure the individual fitness of generators. In the multi-classification, we compare the proposed method with three the state-of-the-art algorithms which are Class-Balanced\cite{cui2019class}, DOS\cite{ando2017deep} and GAMO\cite{mullick2019generative} respectively. All our experiments have been repeated 5 times to mitigate any bias generated due to randomization and the means of the index values are reported.
%Our experiments use GANs as an oversampling method to solve the class imbalance problem. In particular, we only use the samples of minority class to train a GAN, and the well trained GAN will generate data to restore the imbalanced dataset into balanced. The ability of GANs to solve class imbalance problem is reflected by comparing the performance of classifiers trained on imbalanced data and balanced datasets. To ensure that the classifier is fully trained, we perform data augmentation (DA) on the data before and after the balancing, respectively.

%In the experiments, we choose the classic SVM as the classifier. To shorten the training time, we use ThunderSVM \cite{wenthundersvm18} that supports GPU for training. We use WGAN, WGANgp, E-GAN and AGGAN for oversampling and make comparisons, and the same network structures are used for these different methods to achieve a fair comparison. 
%In the experiments using MNIST and Fashion-MNIST datasets, the discriminator consists of three convolutional layers, and the generator consists of three deconvolutional layers. In the experiments of CIFAR-10 and SVHN, the two networks have six layers, respectively. 

\begin{table}[!t]
	\centering
	\begin{tabular}{cccccccccc}
		\toprule
		Dataset&\multicolumn{3}{c}{MNIST} & \multicolumn{3}{c}{Fashion-MNIST} &	\multicolumn{3}{c}{SVHN}\\	  
		\cmidrule{1-1}\cmidrule(lr){2-4} \cmidrule(lr){5-7} \cmidrule(lr){8-10} 
		IR& 10 & 50 &100 & 10 & 50 &100 &10 & 50 &100 \\
		\midrule
		CN &97.93 &94.80 &90.27	&93.53	&86.73	&80.00	&74.65	&58.15&52.00\\
		Os+CN  &98.60	&96.33	&95.53	&94.67	&92.07	&88.13	&87.60	&66.50	&58.90\\
		ACGAN   &99.53	&98.13	&96.60	&\textbf{97.80}	&96.20	&94.13	&90.50	&74.65	&70.55\\
		E-GAN &99.53 &98.33	&97.00	&97.67	&96.13	&94.40	&90.85	&74.90	&75.50 \\
		\textbf{AGGAN} &\textbf{99.60}	&\textbf{98.47}	&\textbf{97.53}	&\textbf{97.80}	&\textbf{96.53}	&\textbf{95.07}	&\textbf{90.95}	&\textbf{81.60}	&\textbf{77.70}\\
		\midrule
		Dataset&\multicolumn{3}{c}{CIFAR} & \multicolumn{3}{c}{CELEBA} &	\multicolumn{3}{c}{LSUN}\\	  
		\cmidrule{1-1}\cmidrule(lr){2-4} \cmidrule(lr){5-7} \cmidrule(lr){8-10} 
		IR& 10 & 50 &100 & 10 & 50 &100 &10 & 50 &100 \\
		\midrule
		CN &72.00 &58.85	&57.65	&71.10 &64.40	&60.80	&81.20	&74.60 &52.00\\
		Os+CN &90.60	&82.75	&80.55	&86.70	&78.85	&77.65	&92.05	&89.95 &64.00
\\
		ACGAN   &92.00	&84.70	&83.65	&91.65	&81.40	&78.60	&95.05	&91.10	&89.50\\
		E-GAN &92.10	&86.55	&85.75	&91.65	&81.75	&80.80	&95.20	&92.40	&91.00\\
		\textbf{AGGAN} &\textbf{94.00}	&\textbf{89.25}	&\textbf{86.25}	&\textbf{92.05} &\textbf{84.20}	&\textbf{81.40}	&\textbf{95.35}	&\textbf{93.10}	&\textbf{91.70}\\
		\bottomrule
	\end{tabular}
	\caption{Accuracy(\%) on imbalanced binary classification with various imbalance ratio.}
	\label{tab:2}
\end{table}

\begin{figure}[!t]
\centering
\includegraphics[width=5in]{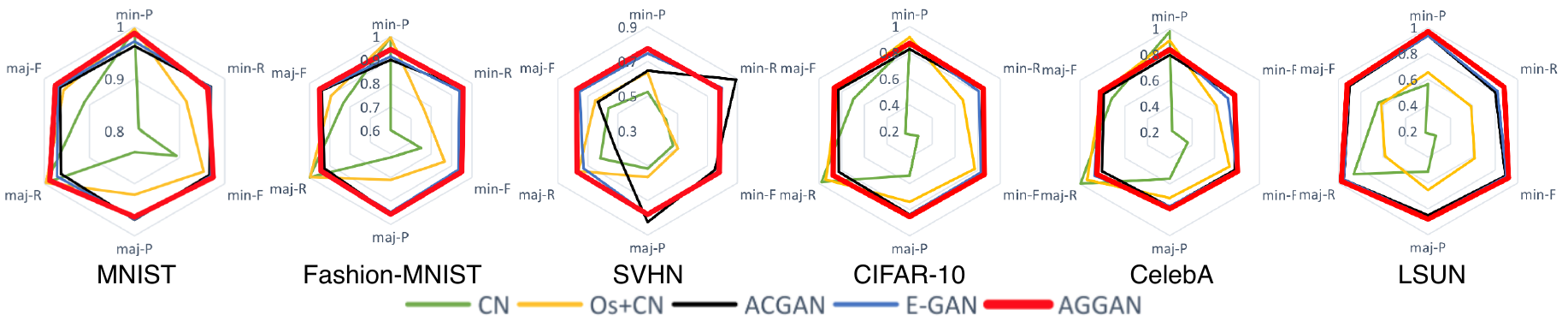}
\caption{Precision, recall, F1-score of majority and minority classes.}
\label{fig_sim3}
\end{figure}
\subsection{Classification Performance}

Table \ref{tab:2} shows the accuracy in binary classification under different imbalance ratios. The experimental results of the original imbalance data and the random oversampled data indicate that data imbalance can significantly affect the performance of the classifier, and simply repeating the minority data could not lead to better performance. The experimental results using GANs for oversampling indecate that GANs can mitigate class imbalance problem. Meanwhile, we can observe that when the IR is low, \eg 10, different GANs perform equally well. However, as the degree of imbalance increasing, the advantages of AGGAN become obvious. When the IR reaches 100, the proposed AGGAN outperforms all other GANs significantly. 
Figure \ref{fig_sim3} shows the three indicators (precision, recall, and F1-score) of the majority and minority classes in the testing dataset under different methods. We can see that for imbalanced data, the recall of the minority class and the precision of the majority class are low. This is because the classifier will judge the data as the majority class as much as possible to improve the accuracy. The proposed method outperforms the other three methods, which is particularly evident in the recalls of minority class. This provides solid evidence that the ability of AGGAN to reconstruct the distribution of the minority class can make more minority data be classified correctly and improve the performance of the classifier significantly. 
Table \ref{tab:4} shows the experimental results on multi-classification. It indicates that AGGAN can still perform well on more complex imbalanced datasets and achieve the state-of-the-art level.
Compared with ACGAN, E-GAN uses the idea of evolution, and our method further combines the methods of genetic and simulated annealing. The experimental results of these three methods fully indicate the effectiveness of the modules of genetic and simulated annealing in AGGAN.
\begin{table}[!t]
    \centering
    \begin{tabular}{c|cccc|ccc}
	    \toprule
		Dataset & CN & ACGAN & EGAN  &\textbf{AGGAN} & CB-loss &DOS & GAMO\\
		\hline
		MNIST  &89.87&90.60&91.41 &\textbf{92.41} &92.24 &90.60 &91.01 \\
		Fashion-MNIST  &79.64&80.51&82.50 &\textbf{83.20} &82.84   &82.74 &83.00   \\
		\bottomrule
	\end{tabular}
	\caption{Accuracy (\%) of imbalanced multi-classification in MNIST and Fashion-MNIST.}
	\label{tab:4}
\end{table}

\begin{figure}
\begin{minipage}{0.55\linewidth}
\centerline{\includegraphics[width=2.5in]{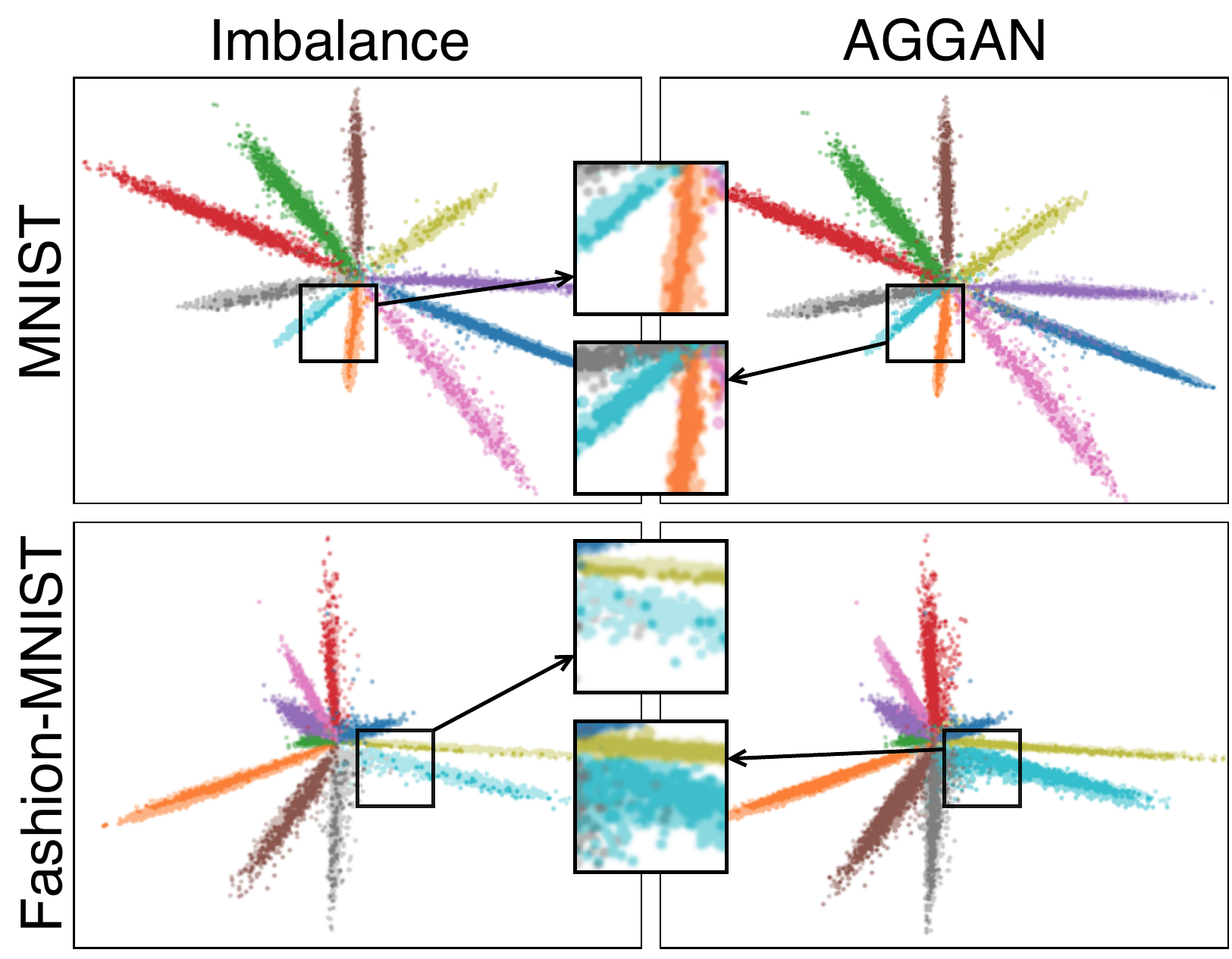}}
 \centerline{(a)}
%\centering
%\includegraphics[width=2.5in]{BMVC/images/Figure4.pdf}
%\caption{Feature visualization for MNIST and Fashion-MNIST.}
%\label{fig_sim5}
\end{minipage}
\hfill
\begin{minipage}{0.45\linewidth}
%  \centering
%\includegraphics[width=2in]{BMVC/images/Figure5.pdf}
\centerline{\includegraphics[width=2in]{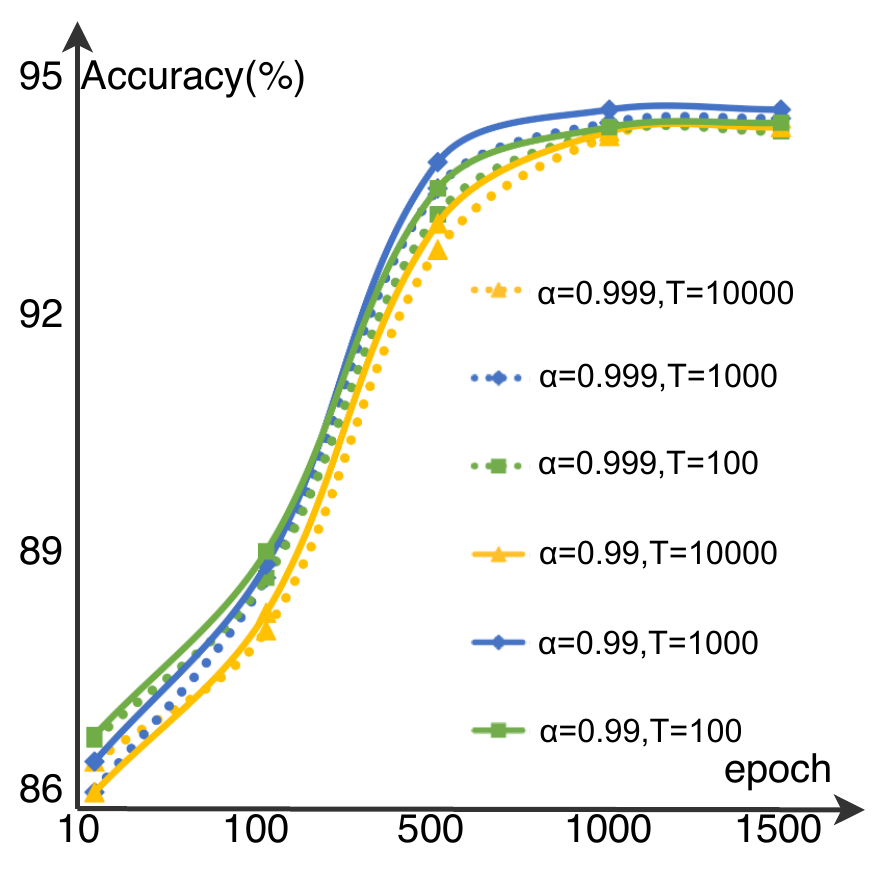}}
 \centerline{(b)}
\end{minipage}
\caption{(a) Feature visualization for MNIST and Fashion-MNIST. (b) Simulated annealing hyper-parameter analysis.}
\label{fig_sim4}
\end{figure}

\subsection{Visualization}
Figure \ref{fig_sim4} (a) we visualize the features of MNIST and Fashion-MNIST datasets before and after using AGGAN to balance the data, respectively. We obtain features by forwarding images to a classifier pre-trained on the original training set, and features with a specific category in each figure are represented in the same color. The test set of each category is represented by translucent dots of corresponding colors. The first column shows the original imbalanced training datasets, in which the training samples of minority class can only cover part of the minority distribution of the test set (which is balanced). As a result, many samples belong to minority class will be misclassified. However, in the second column of Figure \ref{fig_sim4} (a), we can see that after using AGGAN for over-sampling, the minority samples in the training set can almost cover the complete distribution of the minority class in the test set, so the performance of the classifier can be significantly improved. These results indicate that AGGAN can learn realistic distribution from scarce minority samples, and in turn prove the superior performance of AGGAN from the perspective of data distribution.

\subsection{Hyper-parameters analysis}
For a better understanding of the role of the initial temperature $T$ and the annealing coefficient $\alpha$ proposed in AGGAN, we use the CIFAR-10 dataset with IR 10 to show the accuracy and training epochs of the proposed method in Figure \ref{fig_sim4} (b). The search for hyper-parameters are $T\in\{100,1000,10000\}$, $\alpha \in \{0.99,0.999\}$. We have the following observations that at the beginning of training, smaller value of $T$ and $\alpha$ make the model converge faster, while the larger $T$ and $\alpha$ (\eg $T=10000$,$\alpha=0.999$) make the model converge more slowly. However, with an increasing training epoches, the model can achieve higher accuracy finally. These results can indicate that AGGAN is robust to the different values of hyper-parameters.

\section{Conclusion}
In this work, we propose a novel training strategy for GANs, dubbed AGGAN, which aims to reproduce the distributions closest to the ones of the minority classes using limited data samples. Both theoretical analysis and comprehensive experimental studies have shown the robustness and efficacy of our AGGAN.

\bibliography{egbib}
\end{document}